\documentclass[letterpaper, 10 pt, conference]{ieeeconf}  % Comment this line out if you need a4paper
\usepackage{xcolor}
\usepackage{graphicx}
\usepackage{amsmath}
\usepackage{cite}
\usepackage{hyperref}
\usepackage{fontawesome5} % gives you GitHub and many other icons
\IEEEoverridecommandlockouts                              % This command is only needed if 
\title{\LARGE \bf
SADCHER: Scheduling using Attention-based Dynamic Coalitions of
Heterogeneous Robots in Real-Time}

\author{Jakob Bichler$^{1}$, Andreu Matoses Gimenez$^{1}$, Javier Alonso-Mora$^{1}$% <-this % stops a space
\thanks{*This paper has been accepted for publication at the 2025 IEEE Int. Symposium on Multi-Robot \& Multi-Agent Systems (MRS).}
\thanks{$^{1}$Authors are with the Department for Cognitive Robotics, ME, Delft University of Technology, Delft, Netherlands, }}%

\begin{document}

\maketitle
\thispagestyle{empty}
\pagestyle{empty}

%%%%%%%%%%%%%%%%%%%%%%%%%%%%%%%%%%%%%%%%%%%%%%%%%%%%%%%%%%%%%%%%%%%%%%%%%%%%%%%%
\begin{abstract}
We present Sadcher, a real-time task assignment framework for heterogeneous multi-robot teams that incorporates dynamic coalition formation and task precedence constraints. Sadcher is trained through Imitation Learning and combines graph attention and transformers to predict assignment rewards between robots and tasks. Based on the predicted rewards, a relaxed bipartite matching step generates high-quality schedules with feasibility guarantees. We explicitly model robot and task positions, task durations, and robots’ remaining processing times, enabling advanced temporal and spatial reasoning and generalization to environments with different spatiotemporal distributions compared to training. Trained on optimally solved small-scale instances, our method can scale to larger task sets and team sizes. Sadcher outperforms other learning-based and heuristic baselines on randomized, unseen problems for small and medium-sized teams with computation times suitable for real-time operation. We also explore sampling-based variants and evaluate scalability across robot and task counts. In addition, we release our dataset of 250,000 optimal schedules: \href{https://autonomousrobots.nl/paper_websites/sadcher_MRTA/}{\nolinkurl{autonomousrobots.nl/paper_websites/sadcher_MRTA/} \faGithub}
\end{abstract}

%%%%%%%%%%%%%%%%%%%%%%%%%%%%%%%%%%%%%%%%%%%%%%%%%%%%%%%%%%%%%%%%%%%%%%%%%%%%%%%%
\section{INTRODUCTION}
Autonomous multi-robot systems (MRS) are designed for complex, real-world settings, including earthquake disaster response scenarios \cite{zhang_task_2024}, autonomous construction \cite{gosrich_multi-robot_2023}, production assembly processes \cite{gombolay_fast_2018}, or search and rescue missions \cite{ansari_colossi_2024}.  Interest in MRS and multi-robot task assignment (MRTA) has grown rapidly in recent years \cite{chakraa_optimization_2023}. Efficient MRTA algorithms optimize resource usage and minimize operational time \cite{bischoff_multi-robot_2020}. MRS improve performance and enhance system robustness as a multi-robot team is more resilient against individual robot failures and performance bottlenecks \cite{ramachandran_resilience_2019, khamis_multi-robot_2015}. Using sub-teams of robots, i.e., dynamic coalition formation, enables teams to tackle complex tasks that would otherwise be infeasible for a single robot \cite{khamis_multi-robot_2015, quinton_market_2023, babincsak_ant_2023}. In practice, relying on a team of homogeneous robots where each robot possesses all skills can become impractical if task requirements are highly diverse, involving different sensors and actuators \cite{fu_robust_2021}. Instead, using heterogeneous robots brings practical and economic advantages, by leveraging existing specialized robots \cite{muhuri_immigrants_2017} and deploying simpler robots that are more cost-effective to implement and maintain \cite{khamis_multi-robot_2015} and more robust to failures \cite{bischoff_multi-robot_2020}. MRS operate in dynamic environments where sudden changes, new tasks, unexpected task requirements, robot malfunctions, or moving obstacles can occur \cite{chakraa_optimization_2023}. Hence, the ability to adaptively replan in real-time is essential. Modeling precedence constraints, which impose a logical temporal sequence among tasks, further enhances applicability to real-world scenarios \cite{gini_multi-robot_2017} where some tasks depend on the completion of prior tasks. 

% E.g., in autonomous construction, gathering materials must precede assembly \cite{gosrich_multi-robot_2023}. 

Motivated by these challenges this paper proposes Sadcher, a framework for real-time scheduling of heterogeneous multi-robot teams with dynamic coalition formation and precedence constraints. Our main contributions are:
\begin{itemize}
    \item A learning-based model, combining graph attention networks and transformers, which accounts for robot/task positions, task dependencies and durations, robots’ capabilities, and robots' remaining time to complete the current task. This enables advanced spatiotemporal reasoning and generalization.
    \item A dataset of 250,000 optimally solved small-scale problems, usable as demonstrations for imitation learning or to benchmark against optimal schedules.

\end{itemize}
\begin{figure}[t]
    \centering
    \includegraphics[width=1\columnwidth, clip, trim = 135pt 38pt 135pt 0pt] {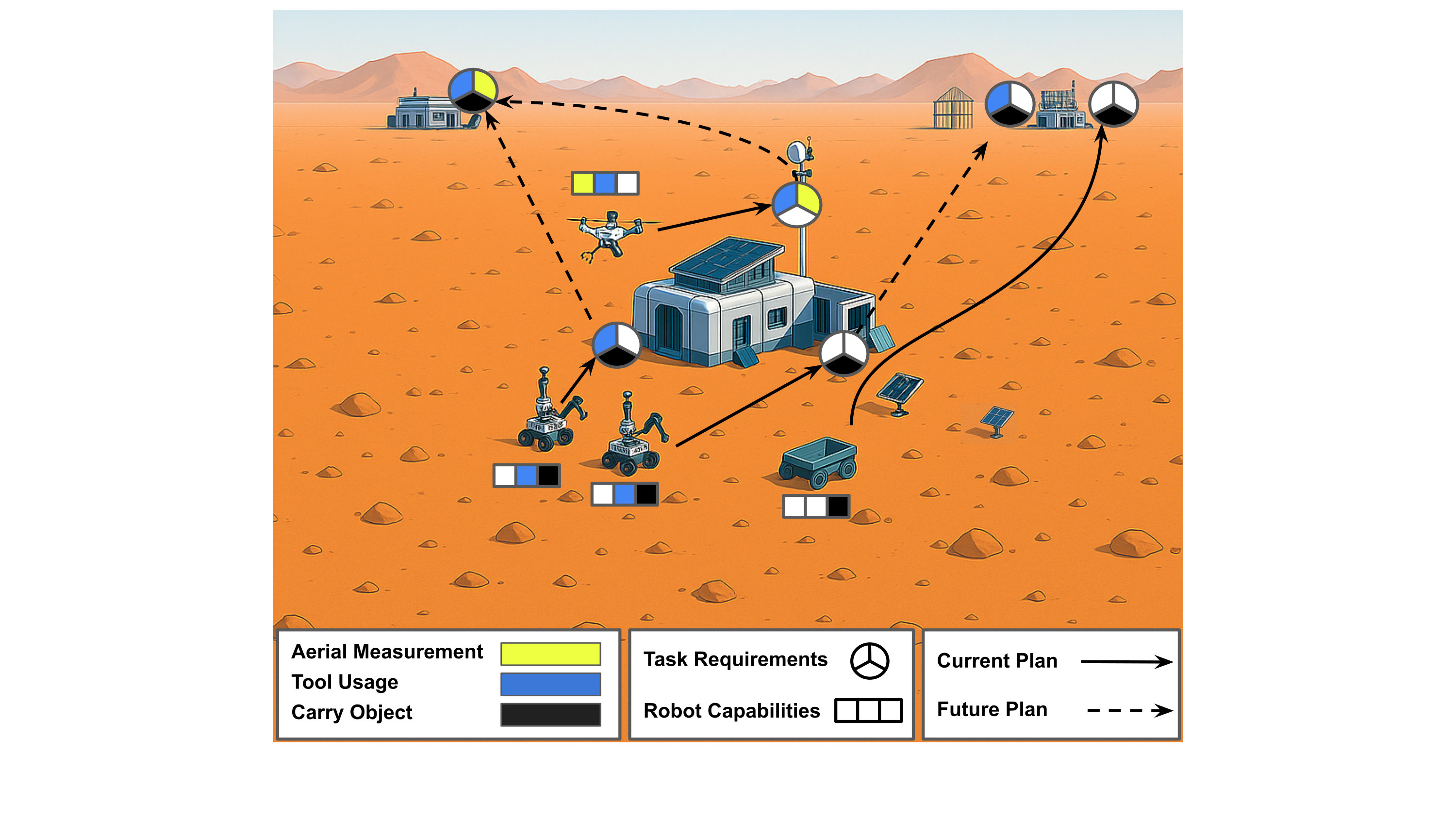} %left lower right upper
    \caption{Illustrative use case of autonomous construction on Mars. Circles represent tasks, color indicates required skills. Robot skills are shown as colored squares. Tasks requiring skills no single robot has (e.g., search for material in top left) must be executed by a synchronized coalition of robots.  }
    \label{fig:problem_instance}
\end{figure}

\section{RELATED WORK}

Following the taxonomy introduced in \cite{gerkey_formal_2004} and extended in \cite{korsah_comprehensive_2013},  MRTA problems can be categorized on 4 axes: (1)~single- or multi-task robots (ST/MT), (2)~single- or multi-robot tasks (SR/MR), (3)~instantaneous or time-extended assignment (IA/TA) where TA incorporates information about future tasks and scheduling decisions, (4)~interdependence of agent-task utilities, i.e. with cross-dependencies (XD), an agent’s utility for a task depends on the tasks assigned to other agents. This work addresses ST-MR-TA-XD settings. 

\subsection{Conventional Methods}
 Mixed Integer Linear Programming (MILP) offers exact solutions for complex ST-MR-TA-XD problems \cite{aswale_heterogeneous_2023}, though its exponential runtime hinders real-time use. The MILP-based CTAS framework \cite{fu_robust_2021} explicitly models risk in agent capability for task decomposition and scheduling. Simpler heterogeneous ST-SR scenarios can be addressed with the Tercio algorithm \cite{gombolay_fast_2018}, which uses an MILP-based task allocator and a polynomial runtime task scheduler.  
Auction algorithms like \cite{ansari_colossi_2024, ansari_cooperative_2020} treat tasks as non-atomic -- tasks execution can be incremental, not requiring coalition formation. \cite{irfan_auction-based_2016} uses auctions to solve heterogeneous ST-MR problems with atomic tasks. 
Genetic Algorithms offer anytime solutions that balance exploration and exploitation: \cite{chakraa_centralized_2023} tackles heterogeneous ST-SR, \cite{arif_robot_2021} focusses on coalition formation of homogeneous robots, while \cite{muhuri_immigrants_2017} can handle heterogeneity and coalition formation. 
Other optimization metaheuristics applied to heterogeneous ST-MR  include Ant Colony Optimization \cite{babincsak_ant_2023} and Particle Swarm Optimization \cite{liu_strength_2023}.
Greedy formulations like \cite{bischoff_multi-robot_2020} employ construction and improvement heuristics to balance runtime and performance.

\subsection{Learning-based Methods}
 Deep learning methods promise fast solution generation and good scalability, by offloading most of the computation to the training phase \cite{prorok_holy_2021}.  Reinforcement Learning (RL) does not require a training dataset, but might spend a lot of time on infeasible solutions \cite{wang_heterogeneous_2022}. RL is used to solve ST-SR problems with mildly heterogeneous robots -- robots differ in efficiency but can all perform any task -- in \cite{paul_learning_2022} and \cite{altundas_learning_2022}.  Other RL methods solve ST-MR problems with dynamic coalition formation, but only for homogeneous robots \cite{deng_learning_2022, dai_dynamic_2024}. Recently, RL has been used to tackle heterogeneous ST-MR problems with dynamic coalition formation in \cite{dai_heterogeneous_2025}. The authors mitigate some of the problems RL faces through a flash-forward mechanism  which allows for decision reordering to avoid deadlocks during training.

 Instead of RL, other methods use Imitation Learning (IL) from optimal solutions during training, which requires a (computationally expensive) expert dataset, but benefits from stable training. \cite{wang_heterogeneous_2022} presents an IL method for mildly heterogeneous robots without coalition formation (ST-SR). Both \cite{gao_collaborative_2023} and \cite{jose_learning_2024} address heterogeneous ST-MR problems with a network predicting task assignment rewards and a bipartite matching algorithm yielding task assignments based on these rewards. In \cite{gao_collaborative_2023}, coalition formation is only considered if a task fails to be completed by a single robot. There are cases in which a lower cost schedule could be obtained by considering coalition formation for all tasks, e.g., two robots are faster at completing the task than one. \cite{jose_learning_2024} improves upon this by always considering coalition formation. Furthermore, they introduce voluntary waiting, which increases performance through enabling better future coalition formation by delaying task assignments. However, \cite{jose_learning_2024} omits locations and durations in their network architecture. This implicitly assumes task durations and travel times to be negligible or to match the training distribution. 

 In this paper, we extend previous IL methods by explicitly modeling robot/task positions, task durations, and robots’ remaining time to complete the current task. This enables advanced spatiotemporal reasoning, e.g., synchronizing robot arrivals and anticipating task readiness and robot availability. Additionally, it supports generalization to environments with unseen spatiotemporal distributions.

\section{PROBLEM STATEMENT}
\label{section:problem_statement}

\textbf{Notation.} Matrices are boldface uppercase (e.g.\ $\mathbf{M}\in\rm I\!R^{n\times m}$), vectors are boldface lowercase (e.g.\ $\mathbf{v}\in\rm I\!R^d$), and scalars are lowercase (e.g.\ $s$).

We model a system of $N$ heterogeneous robots, $M$ tasks, and a set of skills ${\mathcal{S}}$. Each robot is capable of performing a subset of $\mathcal{S}$, and each task requires a subset of $\mathcal{S}$ to be performed at its location for the given task duration. 

The $N$ heterogeneous robots with $S_i \subseteq \mathcal{S}$ distinct skills, are modeled as an undirected graph $\mathcal{G}^{r}=(\mathcal{R}, \mathbf{C})$, where each vertex in $\mathcal{R}=\left\{\mathbf{r}_{i}\right\}^{N}$ is a robot with $d_r$ dimensions. Robot states $\mathbf{r}_{i}=\left[\mathbf{{p_{i}^r}}, t_i^r, a_{i}^{r}, \mathbf{c_{i}^{r}} \right]$  include position $\mathbf{{p_{i}^r}} \in {\rm I\!R ^2}$, remaining duration at the current task $t_i^r$,  the robot's availability $a_i^r \in \{0,1\}$, and the binary capability vector over the global skill set $ \mathbf{c_{i}^{r}} \in \{0,1\}^{|\mathcal{S}|}$.  $\mathbf{C} \in\{0,1\}^{N \times N}$ represents the network connection among the robots. For simplicity, we assume a fully connected graph, so $C_{i,j} = 1 $ $  \forall i,j$, but the model is designed to accept any connected graph as input.

The $M$ tasks and their respective precedence constraints are represented as a directed acyclic graph $\mathcal{G}^{t}=(\mathcal{T}, \mathbf{P})$. Each task is a vertex in $\mathcal{T}=\{\mathbf{t}_{j}\}^{M}$ with  $d_t$ dimensions, and is described by 
 $\mathbf{t}_{j}=\bigl[\mathbf{{p_{j}^t}}, t_j^t,  \mathbf{r_{j}^{t}}, s_{j}^{t}\bigr]$, with position $\mathbf{{p_{i}^r}} \in {\rm I\!R ^2}$, expected duration $t_j^t$, required skills $ \mathbf{r_{j}^{t}} \in \{0,1\}^{|\mathcal{S}|}$ and status $\mathbf{s_{j}^{t}} \in \{0,1\}^{3}$. The status indicates whether tasks are ready, assigned, or incomplete, e.g., $\mathbf{s}_j^t = \left[1,0,1\right] $ represents a task that is ready to be scheduled, currently not assigned, and incomplete. Precedence constraints are encoded in the edges $\mathbf{P}^{M \times M}$, where $P_{i,j}=1$ means the $i$-th task is a predecessor of the $j$-th task. The $j$-th task is only ready to be scheduled if all its preceding tasks have been completed. A task can only commence when all required skills are covered by the dynamically formed coalition of robots assigned to it. This can be denoted as $\mathbf{c_C} \succeq \mathbf{r_j^t}$ where $\mathbf{c_C}$ is the element-wise sum of robot capabilities $\mathbf{c_i^r}$ of assigned robots and $\succeq$ is the element-wise greater-or-equal operator. The tasks require tightly coupled coalitions \cite{arif_robot_2021} - all robots have to be present at the task location for the entire execution duration. Furthermore, we introduce an idle task $t_{M+1}$ that robots can choose to increase overall performance by delaying assignments until a better coalition can be formed.
 
 % The idle task is always ready to be assigned. 

 Robots start at location  $\mathbf{p}_i^{\text{start}}$ and end at  $\mathbf{p}_i^{\text{end}}$. The cost function aims to minimize the makespan, defined as the latest arrival time of any robot at $\mathbf{p}_i^{\text{end}}$ after completing its tasks:
\begin{equation}
\min \max_{i \in \{1, \dots, N\}} \left( t_i^{\text{finish}} + \tau\left( \mathbf{p}_i^{\text{finish}}, \mathbf{p}_i^{\text{end}} \right) \right)
\end{equation}
where $t_i^{\text{finish}}$ is the time robot $i$ finishes its final task, which is computed as the sum of its execution times, idling times, and travel times. $\tau\left( \mathbf{p}_i^{\text{finish}}, \mathbf{p}_i^{\text{end}} \right)$ is the travel time from the location of the last finished task $\mathbf{p}_i^{\text{finish}}$ to the end location $\mathbf{p}_i^{\text{end}}$. Travel times can be estimated using Euclidean distance or path planning algorithms that take obstacles into account.
% For simplicity, we focus on binary skill constraints.
% This formulation is similar to \cite{jose_learning_2024}, but extended by explicitly modeling robot and task positions, task durations, and robots’ remaining time to complete the currently assigned task. 

\section{METHOD}
\label{section:method}

\begin{figure*}[t]
    \centering
    \includegraphics[width=\textwidth]{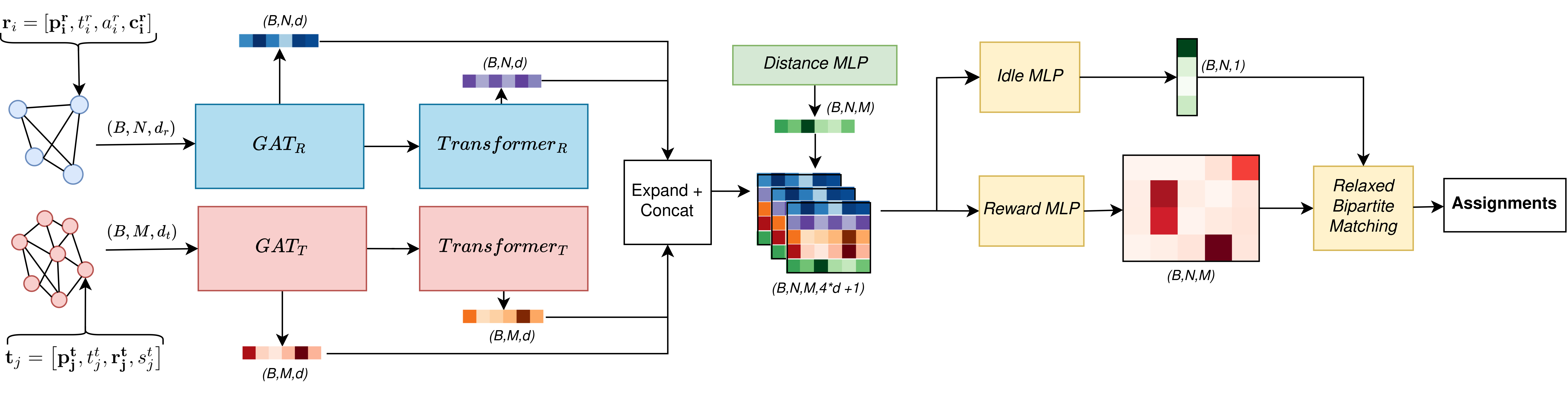}
    \caption{Sadcher architecture overview. Robot and task graphs are processed by graph attention and transformer encoders and concatenated with distance features. The reward matrix is estimated by the Idle and Reward MLPs and final task assignments are extracted using relaxed bipartite matching.  $B$: batch size, $N$: number of robots, $M$: number of tasks, $d_r$: robot input dimension, $d_t$: task input dimension, $d$: latent dimension.}
    \label{fig:architecture}
\end{figure*}

The Sadcher framework consists of a neural network based on attention mechanisms to predict assignment rewards for robots to tasks that is agnostic to the size of the input graphs, i.e., can handle arbitrary numbers of robots and tasks. A relaxed bipartite matching algorithm extracts task assignments based on the predicted reward. During runtime, the method asynchronously recomputes assignments at decision steps, i.e., when robots finish tasks or new tasks are announced.

% we recompute new assignments for available robots. 

%%%%%%%%%%%%%%%%%%%%%%%%%%%%%%%%%%%%%%%%%%%%%%%%%%%%%%%%%%%%%%%%%%%%%%%%%%%%%%%%%%%%%%%%%%%%%%%%%%%%%%%%%%%%%%%%%%%%%%%%%%%%%%%%%%%%%%%%%%%%%%%%%%%%%%%%%%%%%%%%%%%%%%%%%%%%%%%%%%%%%%%%%%%%%%%%%%%%%%%%
\subsection{Network Architecture}

The high-level network structure is depicted in Fig. \ref{fig:architecture} and is similar to \cite{jose_learning_2024}, but extended with a distance multilayer perceptron (MLP) that informs the network about relative distances between robots and tasks and separate heads for predicting rewards for "normal" tasks and the idle action.

The key components of the network are graph attention encoders (GAT) \cite{velickovic_graph_2018}, transformer blocks \cite{vaswani_attention_2017}, and reward MLPs that project latent embeddings into a reward matrix.

\subsubsection{Graph Attention (GAT) Encoder Blocks}

After mapping robot features $\mathbf{r}_i$ and task features $\mathbf{t}_j$ into $d$-dimensional embeddings, the embedded robot and task features are processed by separate GATs to capture local information-rich latent representations of the input graphs. GATs process a set of node features, incorporating information from neighboring nodes based on an adjacency matrix. While the robot features are processed as a fully connected graph, assuming all-to-all attention, the task GAT leverages the encoded precedence constraints in the adjacency matrix to understand the temporal task logic. A single head of the GAT computes attention weights $\alpha_{i,j}$ between node $i$ and its neighbors $j$, based on a projected feature vector $\mathbf{h}' = \mathbf{hW}^h$ (where $\mathbf{h}$ is the input node feature and $\mathbf{W}^h$ is a learned weight matrix):

\begin{equation}
\alpha_{i,j} = \frac{\exp(\text{LeakyReLU}(a([\mathbf{h}'_i || \mathbf{h}'_j])))}{\sum_{k \in \mathcal{N}i \cup {i}} \exp(\text{LeakyReLU}(a([\mathbf{h}'_i || \mathbf{h}'_k])))}
\end{equation}

\noindent
here, $a$ is a learnable linear transformation, $||$ denotes concatenation and $\mathcal{N}_i$ is the set of neighbors of node $i$. A Leaky ReLU \cite{xu_empirical_2015} in combination with a softmax function over the neighbors of node $i$ yields the final $\alpha_{i,j}$. The resulting $\alpha_{i,j}$ represent the relative importance of node $j$ to node $i$, enabling context-aware feature propagation. Spatiotemporally related tasks or robots with complementary skills will attend more strongly to each other. The output $\mathbf{h}^{GAT}_i$ for a single head at node $i$  is a sum of a self-loop contribution and the transformed neighbor contributions:

\begin{equation}
   \mathbf{h}^{\text{GAT}}_i = \alpha_{i,i} \mathbf{h}'_i + \text{LeakyReLU}\left(\sum_{j \in \mathcal{N}_i, j \neq i} \alpha_{i,j} \mathbf{h}'_j\right) 
\end{equation}

In the GAT encoder blocks, we apply multi-head GAT, concatenating the outputs of $Z_{\text{GAT}}$ independent heads and applying residual connections and layer normalization. The GAT encoder consist of $L_{{\text{GAT}}}$ such layers and outputs $\mathbf{h}^{{\text{GAT\_R}}}$ for robots and  $\mathbf{h}^{{\text{GAT\_T}}}$ for tasks respectively.

\subsubsection{Transformer Encoder Blocks}
Following the GAT encoders, the representations $\mathbf{h}^{{\text{GAT\_R}}}$ and $\mathbf{h}^{{\text{GAT\_T}}}$ are processed by independent transformer encoders, to build the global context of robots and tasks. Each transformer block applies multi-head self-attention (MHA)\cite{vaswani_attention_2017} on the input $\mathbf{h}$:
% followed by layer normalization, MLP, and another layer normalization, with residual connections 
% . Multi-head self-attention transforms input queries, keys, and values ($\mathbf{Q}, \mathbf{K}, \mathbf{V}$) derived from the input representation $\mathbf{h}$ via linear projections $\mathbf{W}^Q, \mathbf{W}^K, \mathbf{W}^V$ : 
\begin{equation}
    \alpha_z = \text{Softmax}\left( \frac{(\mathbf{W}^Q_z \mathbf{h}) (\mathbf{W}^K_z \mathbf{h})^\top}{\sqrt{d}} \right) (\mathbf{W}^V_z \mathbf{h})
\end{equation}
\begin{equation}
    {\text{MHA}}(\mathbf{h}) = \!\bigl(\alpha_{1}||\alpha_{2}||\ldots||\alpha_{Z}\bigr)\mathbf{W}^{O}
\end{equation}

\noindent
where $d$ is the key dimensionality. MHA computes this operation in parallel for $Z_{T}$ heads to generate the final outputs $\mathbf{h}^{\text{T\_R}}$ for robots and  $\mathbf{h}^{\text{T\_T}}$ for tasks respectively.

\subsubsection{Reward Prediction}
The normalized relative distances between robot $i$ and task $j$ are passed through the distance head $\text{MLP}_\text{D}$ to compute the distance feature $d_{i,j}$:
\begin{equation}
    d_{i,j} = \text{MLP}_\text{D}\left(\text{Normalize}(\|\mathbf{p}^R_i - \mathbf{p}^T_j\|_2)\right)
\end{equation}

\noindent
While task and robot positions are part of the raw input features in $\mathcal{G}^{r}$ and $\mathcal{G}^{t}$, this explicit distance term provides the network with direct access to spatial proximity.

We construct feature vectors $\mathbf{f}_{i,j}$ for each robot-task pair by concatenating the local (GAT) and global (transformer) representation of robot $i$ and task $j$ with  the distance term $d_{i,j}$, so  $\mathbf{f}_{i,j}\in \rm I\!R^{4 \times d_k + 1}$:

\begin{equation}
    \mathbf{f}_{i,j} = \mathbf{h}^{\text{GAT\_R}}_i \; \Vert \; \mathbf{h}^{\text{GAT\_T}}_j \; \Vert \; \mathbf{h}^{\text{T\_R}}_i \; \Vert \; \mathbf{h}^{\text{T\_T}}_j \; \Vert \; d_{i,j}
\end{equation}

\noindent
This information-rich representation is then passed through the reward head $\text{MLP}_\text{R}$ to compute the task assignment reward $R_{i,j}$ to assign robot $i$ to task $j$. The idle reward $R_{i}^{\text{IDLE}}$ is computed by passing $\mathbf{f}_{i,j}$ to the idle head $\text{MLP}_\text{I}$ and summing the outputs across all tasks for each robot $i$: 
\begin{equation}
    R_{i,j}^{\text{task}} = \text{MLP}_\text{R} \left( f_{i,j} \right), \qquad
    R^{\text{IDLE}}_{i}= \sum_{j=1}^{M}\text{MLP}_{\text{I}}\!\left(f_{i,j}\right)
\end{equation}

\noindent
$\text{MLP}_\text{I}$ can be understood as learning per-task signals that encourage a robot to wait when short-term idling is advantageous (e.g., a nearby task will become ready soon). The final predicted reward $\mathbf{R}$ contains the task rewards $R_{i,j}^{\text{task}}$ for all pairs of robots $i$ and tasks $j$, concatenated with the idle rewards $R_i^{\text{IDLE}}$ for each robot $i$, so $\mathbf{R} \in \rm I\!R^{N \times (M+1)} $.
%%%%%%%%%%%%%%%%%%%%%%%%%%%%%%%%%%%%%%%%%%%%%%%%%%%%%%%%%%%%%%%%%%%%%%%%%%%%%%%%%%%%%%%%%%%%%%%%%%%%%%%%%%%%%%%%%%%%%%%%%%%%%%%%%%%%%%%%%%%%%%%%%%%%%%%%%%%%%%%%%%%%%%%%%%%%%%%%%%%%%%%%%%%%%%%%%%%%%%%%

\subsection{Task Assignment through Bipartite Matching}
The final reward $\mathbf{R} $ can be interpreted as the edge rewards between robots $\mathcal{R}$ and tasks $\mathcal{T}$ at a given timestep, encoding the full complexity of the current problem state. To extract task assignments at this timestep, we employ a relaxed bipartite matching formulation (no strict one-to-one matching). The constraints ensure valid assignments: (\ref{eq:max_1_task_per_robot}) prevents robots from being assigned more than one task, (\ref{eq:capability_must_match}) guarantees that each task’s required skills are fully covered by the assigned coalition using the element-wise inequality $\succeq$, and (\ref{eq:zero_assignment_for_unready_task_unavaiable_robots}) enforces that only idle robots and ready tasks are matched. The bipartite matching finds the optimal assignment matrix $\mathbf{A^*} \in \rm I\!R^{N \times (M+1)} $ that maximizes the selected edge reward encoded in  $\mathbf{R}$:
\begin{equation}
    \mathbf{A}^* = \arg\max_{\mathbf{A}} \sum_{i,j} A_{i,j} R_{i,j}
\end{equation}
subject to:
\begin{align}
    \sum_{j = 0}^{M+1} A_{i,j} &\leq 1, && \forall i \in \mathcal{R} \label{eq:max_1_task_per_robot} \\
        \sum_{i=0}^{N} A_{i,j}\,\mathbf{c}^r_{i} &\succeq \mathbf{c}^t_{j}, && \forall j \in M \label{eq:capability_must_match}\\
    A_{i,j} &= 0, && \forall i,j : i \notin \mathcal{R}_{\text{idle}} \lor j \notin \mathcal{T}_{\text{ready}} \label{eq:zero_assignment_for_unready_task_unavaiable_robots}
\end{align}

This formulation prevents deadlocks since no coalition can be assigned to a task that it cannot execute. However, it allows for redundant assignments, so after computing $\mathbf{A}^*$, we remove robots that do not contribute unique required skills, starting from the robot with the highest travel time to the task.
Additionally, we implement a pre-moving strategy: If robot $r_i$ is assigned the idle task $t_{M+1}$, it moves towards the task with the highest reward $t_{\mathrm{highest}}^i = \arg\max_{1 \le j \le M} R_{i,j}$, without being formally assigned to it. This does not concatenate the tasks into a fixed schedule for the robot, since assignments are recomputed at decision steps. The robot is likely to be assigned to $t_{\mathrm{highest}}^i$ at the next decision step, so pre-moving can reduce the delay to task start if $r_i$ would have been the last coalition member to arrive at $t_{\mathrm{highest}}^i$.

%%%%%%%%%%%%%%%%%%%%%%%%%%%%%%%%%%%%%%%%%%%%%%%%%%%%%%%%%%%%%%%%%%%%%%%%%%%%%%%%%%%%%%%%%%%%%%%%%%%%%%%%%%%%%%%%%%%%%%%%%%%%%%%%%%%%%%%%%%%%%%%%%%%%%%%%%%%%%%%%%%%%%%%%%%%%%%%%%%%%%%%%%%%%%%%%%%%%%%%%

%%%%%%%%%%%%%%%%%%%%%%%%%%%%%%%%%%%%%%%%%%%%%%%%%%%%%%%%%%%%%%%%%%%%%%%%%%%%%%%%%%%%%%%%%%%%%%%%%%%%%%%%%%%%%%%%%%%%%%%%%%%%%%%%%%%%%%%%%%%%%%%%%%%%%%%%%%%%%%%%%%%%%%%%%%%%%%%%%%%%%%%%%%%%%%%%%%%%%%%%
\subsection{Training through Imitation Learning}
\subsubsection{Training Data Generation}
We generate 250,000 small-scale problem instances (8 tasks, 3 robots, 3 skills) with fully randomized configurations: each robot is assigned 1–3 skills, each task requires 1–3 skills, task locations and robot start/end depot lie in $[0,100] \times [0,100] \subset \rm I\!R^2$ and are randomly sampled. Execution times are drawn uniformly from $[50,100]$, precedence constraints are acyclic and generated between random task pairs, and robot travel speed is assumed to be 1 unit per timestep. To solve these scenarios optimally, we extend the exact MILP formulation of \cite{aswale_heterogeneous_2023} with precedence constraints. Due to its exponential time complexity, both in the number of robots and tasks, only small instances can be solved in a reasonable time to generate a training dataset. We omit modelling of stochastic travel times, as our framework handles deviations via real-time replanning and does not need conservative safety margins.

%%%%%%%%%%%%%%%%%%%%%%%%%%%%%%%%%%%%%%%%%%%%%%%%%%%%%%%%%%%%%%%%%%%%%%%%%%%%%%%%%%%%%%%%%%%%%%%%%%%%%%%%%%%%%%%%%%%%%%%%%%%%%%%%%%%%%%%%%%%%%%%%%%%%%%%%%%%%%%%%%%%%%%%%%%%%%%%%%%%%%%%%%%%%%%%%%%%%%%%%

\subsubsection{Optimal Reward Extraction}
To train the network to imitate the optimal behavior, we extract ``ground-truth'' reward matrices $\mathbf{O_k} \in \rm I\!R^{N \times (M+1)}$. The optimal schedules are sliced into $K$ decision points $T_k^{\text{dec}}$, corresponding to timesteps when a task finishes and the robots require reassignment. At each $T_k^{\text{dec}}$ the optimal reward is calculated based on the time difference between $T_k^{\text{dec}}$  and the finish time of task $j$ with discount factor $\gamma \in (0,1]$:
\begin{equation}
    {O_{k,i,j}} = \gamma^{ \bigl( T_j^{\text{finish}} - T_k^{\text{dec}} \bigr)} o_{k,i,j}
\end{equation} 

\noindent
where $o_{k,i,j}=1$ if robot $i$ is assigned to task $j$ in the optimal solution and the decision point occurs before the task’s start time ($T_k^{\text{dec}} < T_{i,j,\text{start}}$); otherwise, $o_{k,i,j}=0$.

% where $o_{k,i,j}$ is 1 if robot $i$ executes task $j$ in the optimal solution and the decision point precedes the task start time:

% \begin{equation}
% o_{k,i,j} = \begin{cases}
% 1, & \text{if } T_k^{\text{dec}} < T_{i,j,\text{start}}  \ \land \ r_i \ \text{executes} \ t_j   \\
% 0, & \text{otherwise}
% \end{cases}
% \end{equation}

We handle the idle action \(t_{M+1}\) in the same way: If the time between a robot’s last finish time and its next start time exceeds the travel time between the corresponding tasks, we treat this interval as an explicit idle assignment in the optimal schedule and compute its reward using the above formulas.

By design, this reward encoding captures the optimal decision logic: The next selected task will have the highest reward, with decreasing rewards for later tasks. Given the sequence of optimal rewards $\mathbf{O_k}$ over all decision steps $T_k^{\text{dec}} $, the bipartite matching algorithm outputs the exact solution.

%%%%%%%%%%%%%%%%%%%%%%%%%%%%%%%%%%%%%%%%%%%%%%%%%%%%%%%%%%%%%%%%%%%%%%%%%%%%%%%%%%%%%%%%%%%%%%%%%%%%%%%%%%%%%%%%%%%%%%%%%%%%%%%%%%%%%%%%%%%%%%%%%%%%%%%%%%%%%%%%%%%%%%%%%%%%%%%%%%%%%%%%%%%%%%%%%%%%%%%%

\subsubsection{Training Details }
We modify the loss $\mathcal{L}$ from \cite{jose_learning_2024} by applying the inverse mask ($1- \mathbf{X_k}$) to the second term: 
\begin{equation}
\label{eq:LVWSLOSS}
\mathcal{L} = \|\mathbf{X_k} \circ (\mathbf{R_k} - \mathbf{O_k})\|_1 + \lambda \|(1 - \mathbf{X_k}) \circ (\mathbf{R_k} - \mathbf{O_k})\|_1
\end{equation}

\noindent
where $\circ$ denotes the element-wise product operator, $\mathbf{O_k}$ is the optimal reward, $\mathbf{R_k}$ is the predicted reward and $\mathbf{X_k} \in \rm I\!R^{N \times (M+1)}$ is a feasibility mask with $X_{i,j} = 1 $ if robot $i$ is available and task $j$ is ready, else $X_{i,j} = 0$. The first term encourages accurate prediction of feasible rewards, while the second discourages high values for infeasible ones. $\lambda$ balances the two terms: Intuitively, accurate feasible predictions are more important than suppressing infeasible ones, as the bipartite matching will select high reward tasks. We use the ADAM optimizer \cite{kingma_adam_2017} to train the network.
%%%%%%%%%%%%%%%%%%%%%%%%%%%%%%%%%%%%%%%%%%%%%%%%%%%%%%%%%%%%%%%%%%%%%%%%%%%%%%%%%%%%%%%%%%%%%%%%%%%%%%%%%%%%%%%%%%%%%%%%%%%%%%%%%%%%%%%%%%%%%%%%%%%%%%%%%%%%%%%%%%%%%%%%%%%%%%%%%%%%%%%%%%%%%%%%%%%%%%%%

\begin{figure*}[t!]
    \centering
    \includegraphics[width=\linewidth, trim=0pt 0 0pt 0, clip]{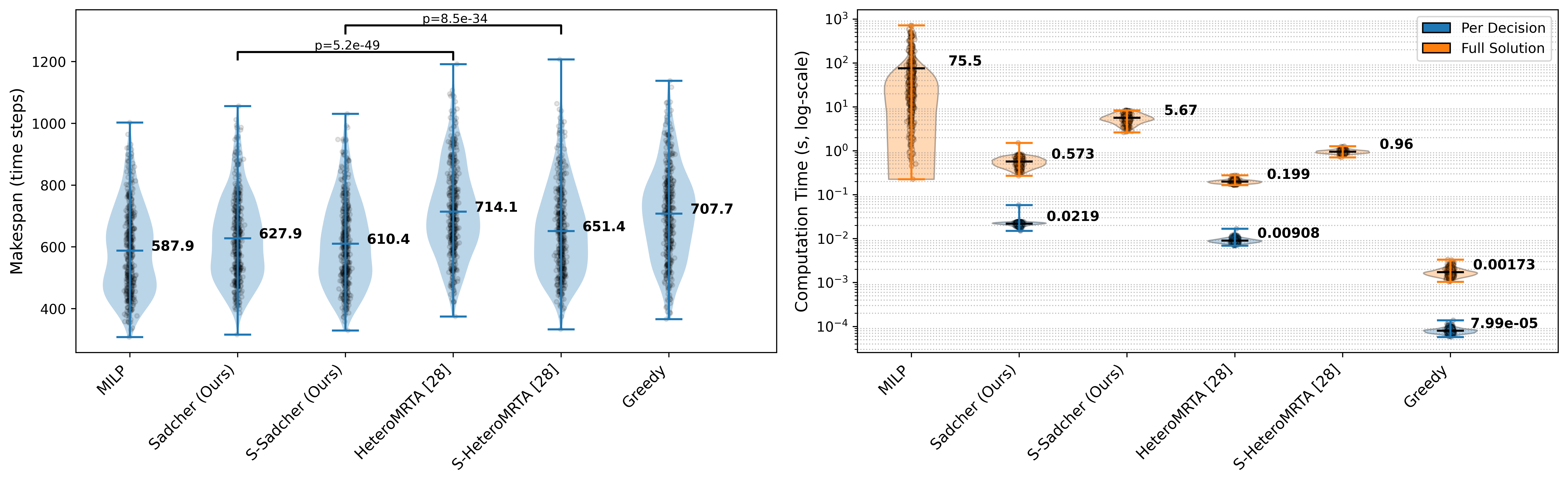}
    \caption{Comparison on 500 unseen, randomized problem instances (8 tasks, 3 robots, 3 precedence constraints) for makespan (left), and computation time (right). Lower means better performance. Wilcoxon significance levels are annotated for Sadcher compared to HeteroMRTA variants. All other pairwise differences are statistically significant ($p < 0.05$), except between S-HeteroMRTA
    and Greedy ($p = 0.21$). For algorithms requiring full solution construction, total computation time is reported; for methods returning instantaneous assignments, both time per decision and total time are shown.}
    \label{fig:violin_plot}
\end{figure*}

\section{EXPERIMENTS AND RESULTS}

\begin{figure}
    \centering
    \includegraphics[width=\linewidth, trim=5pt 0 5pt 0, clip]{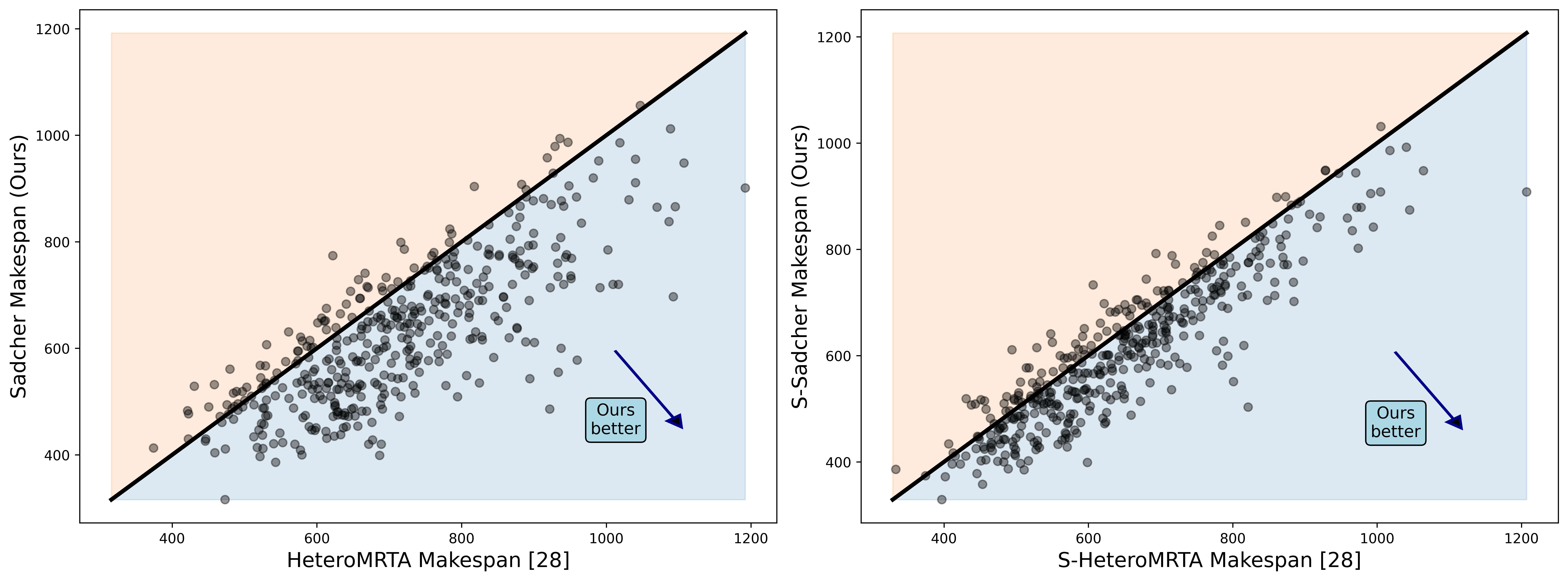}
\caption{Pairwise makespan comparison of HeteroMRTA vs.\ Sadcher (left) and S-HeteroMRTA vs.\ S-Sadcher (right). Each point is one solved instance; points below the diagonal indicate better performance by (S-)Sadcher.}
    \label{fig:1v1}
\end{figure}

We evaluate makespan and computation time metrics for six algorithms, averaged over 500 unseen problem instances with randomized task locations/durations, skill requirements, robot capabilities, and precedence constraints. 
Experiments are conducted on a consumer machine with an AMD Ryzen 7 4800H CPU and NVIDIA 
GeForce GTX 1650 GPU.

\subsection{Compared Algorithms}
\subsubsection{Baselines}
There exist few algorithms in literature that address heterogeneous ST-MR-TA-XD with precedence constraints, often without weights or datasets \cite{jose_learning_2024}. We chose one available algorithm for each of the approaches commonly followed (optimization, learning, heuristic search): \\
(1) An MILP formulation based on \cite{aswale_heterogeneous_2023}, adding precedence constraints and omitting stochastic travel times, which provides optimal solutions with formal guarantees. 
A decentralized RL framework HeteroMRTA \cite{dai_heterogeneous_2025}, adapted for precedence constraints by masking out tasks that have incomplete predecessors during action selection. We compare against (2), the single solution variant HeteroMRTA, where agents choose the highest-probability task at decision steps, and (3), the sampling variant S-HeteroMRTA (Boltzmann weighted-random action selection) which returns the best makespan solution across 10 runs per instance.
We also implement and compare (4), a greedy heuristic that assigns robots to tasks based on reducing the remaining skill requirements the most and breaking ties based on travel time (shortest first).

\subsubsection{Sadcher Variants}
(5) The Sadcher framework predicts robot-task rewards deterministically as described in \mbox{Section~\ref{section:method}}. We also benchmark (6), a S-Sadcher variant, which samples reward matrices from a normal distribution centered around the deterministic output then used by the bipartite matching. This introduces stochastic variations in the schedules. As for S-HeteroMRTA we run this process 10 times per instance and select the best-performing rollout.

\subsection{Training-Domain Evaluation}

We evaluate the algorithms on 500 randomized problem instances of the training domain size (8 tasks, 3 robots, 3 precedence constraints). Results are shown in Fig. \ref{fig:violin_plot} and \ref{fig:1v1}. 

\subsubsection{Makespan}
The MILP formulation provides optimal makespans, establishing a baseline for comparing the average relative gaps of other methods. S-Sadcher (gap: 3.8\%) and Sadcher (gap: 6.8\%) are the best-performing non-optimal algorithms. HeteroMRTA performs worst (gap: 21.5\%), but sampling reduces the optimality gap to 10.8\%, leveraging its RL policy, which follows a sampling strategy during training. In the pairwise comparison in Fig. \ref{fig:1v1}, Sadcher achieves a lower makespan for 403 of 500 instances (80.6\%, binomial test: $\text{p} \approx 2 \times 10^{-45}$). S-Sadcher outperforms S-HeteroMRTA on 389 of 500 instances (77.8\%, binomial test: $\text{p} \approx 3 \times 10^{-37}$). Greedy reaches an average gap of 20.4\%.

% \subsubsection{Travel Distance}
% Although makespan is the primary optimization target, we also report travel distance (in simulation units) as an auxiliary indicator of scheduling efficiency. MILP yields the shortest average travel routes (533.9 units) followed by S-Sadcher (551.9~units), Sadcher (577.8~units), and Greedy (641.6~units). Interestingly, since the greedy formulation is locally biased towards selecting the closest tasks, it finds the lowest travel distance for one instance. In general, low makespans correlate with low travel distances, but this relationship is not strict.  Due to the complex synchronization of robot routes and coalition formation, some schedules can have a short makespan but longer travel distances, or vice versa. S-HeteroMRTA reaches 649.9~units and HeteroMRTA 681.9~units.  

\subsubsection{Computation Time}
For dynamic scenarios with real-time requirements, the time per assignment decision ($t_\text{dec}$) is crucial.
MILP cannot compute instantaneous assignments, but only globally optimal schedules. S-HeteroMRTA and S-Sadcher roll out the full scenario to select the best assignments. Therefore, these three algorithms, do not yield a time per decision, but only for full solution construction ($t_\text{full}$). 
Due to its simplicity, the greedy algorithm computes the fastest ($t_\text{dec}$: 0.080~ms; $t_\text{full}$: 1.7~ms). HeteroMRTA ($t_\text{dec}$: 9.1~ms; $t_\text{full}$: 0.20~s) is faster than Sadcher ($t_\text{dec}$: 22~ms; $t_\text{full}$: 0.57~s), which needs to solve the relatively expensive bipartite matching for each decision. S-HeteroMRTA computes full solutions in 0.96~s, S-Sadcher in 5.7~s, and MILP in 76~s. In the worst case, MILP takes up to 12 minutes, rendering it infeasible for real-time applications, even on small problems.

\subsubsection{Precedence Constraints}
The Sadcher model demonstrates an understanding of task dependencies by prioritizing the assignment of predecessor tasks. This improves performance by unlocking successors earlier and enabling better global schedules.  On average, the model assigns ready predecessor tasks approximately 1.7 times more frequently compared to the baselines. (S-) HeteroMRTA and Greedy cannot make this informed decision, but selecting tasks with incomplete predecessors is prevented through masking. 

% from the pool of ready tasks (including tasks with and without successors) than a strategy that assigns tasks uniformly from the same pool. 

\subsection{Out-of-Domain Generalization}

\begin{figure}
    \centering
    \includegraphics[width=\linewidth]{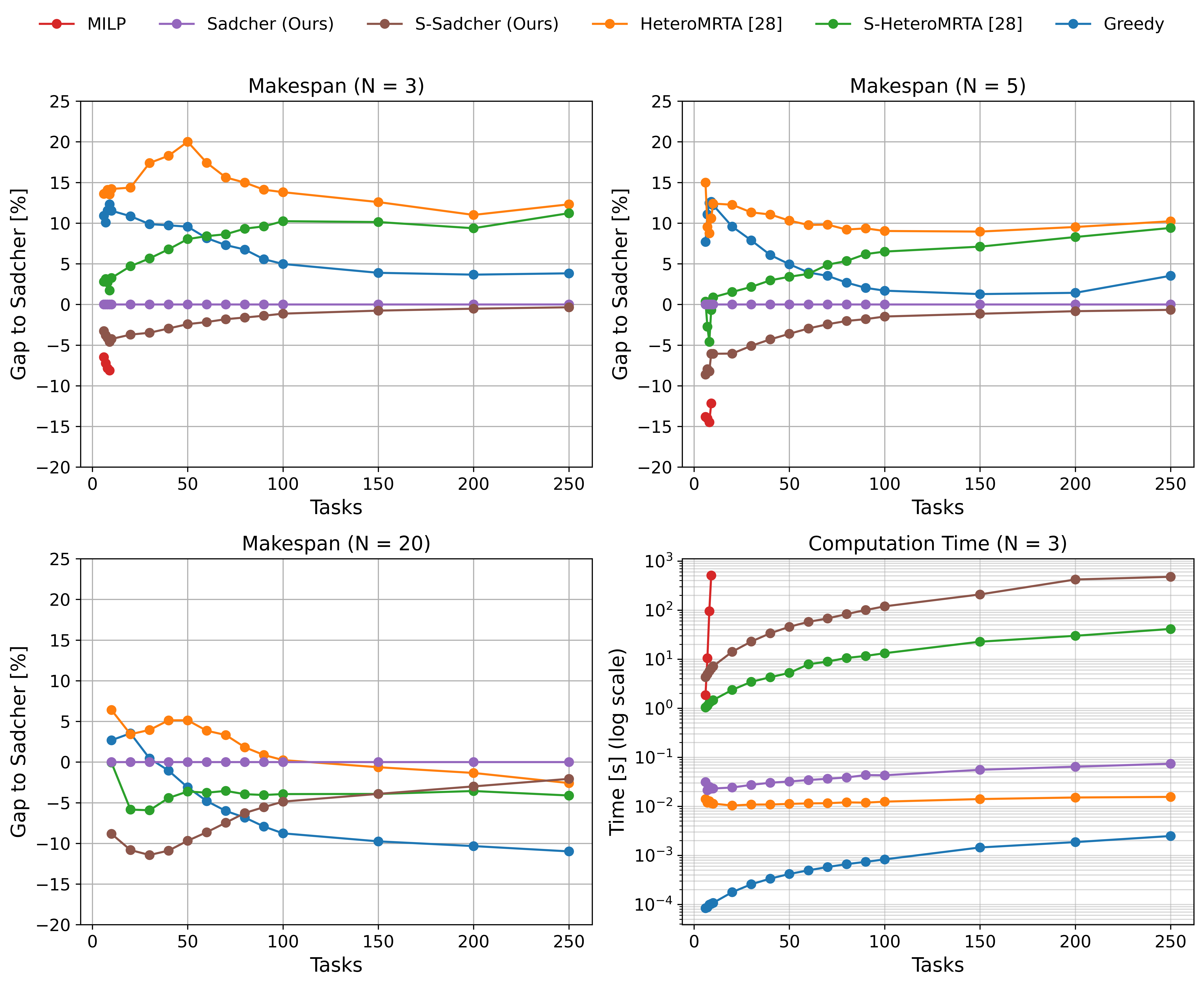}
    \caption{Relative makespan gap to Sadcher for 3, 5, and 20 robots (top left to bottom left). Bottom right: Computation time for 3 robots (for algorithms requiring full solution construction (Sample-HeteroMRTA, Sample-Sadcher, MILP), total computation time is reported, for methods returning instantaneous assignments (HeteroMRTA, Sadcher, Greedy), time per decision is reported). Task counts $M \in [6, 250]$, with $S = 3$ skills and $M/5$ precedence constraints. Each point shows the mean over 100 runs.}
    \label{fig:scaling_grid}
\end{figure}
To evaluate generalization, we scale the number of robots $N \in \{3, 5, 20\}$, and the task number $M \in [6, 250]$ (see Fig.~\ref{fig:scaling_grid}) and compare the makespan gap relative to Sadcher (trained on $N{=}3$, $M{=}8$). With a 1-hour cutoff, the MILP solver fails to find solutions beyond 10 tasks and 7 robots within this limit. For smaller problem sizes, it finds the optimal makespans, outperforming Sadcher by 6-16\%.

For $N{=}3$ robots, S-Sadcher and Sadcher are the strongest non-optimal methods across all $M$. S-HeteroMRTA outperforms Greedy for $M{\leq}60$ and HeteroMRTA finds the highest makespans overall. Although Sadcher's relative performance is best for small $M$, it outperforms Greedy by more than 4\% and (S-) HeteroMRTA by more than 9\% for $M{=}250$. 

For $N{=}5$ robots, S-Sadcher remains the best learning-based method across all $M$. S-HeteroMRTA performs better than Sadcher for $M {\leq} 9$, but is surpassed by Sadcher beyond that, and by Greedy for $M{\geq}70$. HeteroMRTA outperforms Greedy for $7 {\leq} M {\leq} 10$. Greedy reaches a 2\% gap for $M {=}200$.

For $N{=}20$ robots, relative performance changes significantly, with degradation of the learning-based methods: S-Sadcher is the best-performing method only for $M {\leq} 70$,  beyond that, Greedy becomes superior. S-HeteroMRTA consistently beats Sadcher, yet only outperforms Greedy for $M {\leq} 50$. HeteroMRTA surpasses Sadcher for $M {\geq} 150$, and the performance gap for $M {\leq} 100$ between the two is smaller compared to scenarios with fewer robots.

Overall, Sadcher excels on smaller robot teams across all task counts, but its performance decreases with more robots. We hypothesize that increasing the number of tasks while keeping the number of robots fixed is similar to solving multiple smaller subproblems sequentially, where local scheduling rules learned during training remain effective. On the other hand, larger teams require different local scheduling strategies that diverge from the distribution Sadcher has seen during training.
% HeteroMRTA is trained with high agent-to-task ratios, enabling it to perform relatively better on problem instances with a large number of agents and a low to moderate number of tasks. 
For high task counts, the greedy algorithm - especially in combination with bigger robot teams - starts beating the learning-based methods.
Sampling-based variants (S-Sadcher, S-HeteroMRTA) have a higher impact on smaller problems, where the smaller solution space makes rollouts more likely to yield improvements.

Greedy computes fastest, delivering near-instantaneous decisions ($\leq$3~ms). The computation time of HeteroMRTA is minimally affected by scaling ($\leq$20~ms), while Sadcher is slower and scales worse due to the bipartite matching step ($\leq$ 80~ms per decision). The sampling-based variants require significantly longer computation (up to 450~s for S-Sadcher and  40~s for S-HeteroMRTA for $M {=}250$), which makes them impractical for online computation on large problems.

% For $N{=}7$ robots, S-Sadcher remains the best learning-based method for $M{\leq}150$. S-HeteroMRTA beats Sadcher for $M{\leq}40$. However, at higher task counts the performance of all learning-based methods declines: Greedy outperforms HeteroMRTA for $M{\geq} 30$, S-HeteroMRTA at $M{\geq}60$, Sadcher at $M{\geq}100$, and S-Sadcher at $M{\geq}200$.

\section{CONCLUSION}
In this work, we proposed Sadcher - an IL framework to address real-time task assignment for heterogeneous multi-robot teams, incorporating dynamic coalition formation and precedence constraints. Reward prediction with relaxed bipartite matching yields strong performance with feasibility guarantees.
Sadcher outperforms RL-based and heuristic baselines in makespan across small to medium-sized robot teams and a wide range of task counts. For bigger teams, the advantage is lost due to lack of demonstrations. Sadcher can generate assignments in real-time across all tested problem sizes, but the sampling variant S-Sadcher is only real-time for smaller problems. Sadcher relies on a large (computationally expensive) dataset of expert demonstrations for training.

Future work will explore fine-tuning IL policies with RL, which could increase performance on larger problem instances where expert solutions are very expensive or infeasible to obtain.
Additionally, extending the dataset with sub-optimal demonstrations for bigger problem instances could improve scalability. 
% Finally, the field would benefit from a standardized benchmark dataset. While our 250,000-instance dataset targets complex MRTA, expanding it with simpler, larger, and multi-objective scenarios would allow evaluation across a wider range of methods.

%%%%%%%%%%%%%%%%%%%%%%%%%%%%%%%%%%%%%%%%%%%%%%%%%%%%%%%%%%%%%%%%%%%%%%%%%%%%%%%%
\clearpage

\section*{Acknowledgments}
This project has received funding from the European Union through ERC, INTERACT, under Grant 101041863. Views and opinions expressed are, however, those of the author(s) only and do not necessarily reflect those of the European Union. Neither the European Union nor the granting authority can be held responsible for them.

\bibliography{pruned}

\begin{thebibliography}{10}
\providecommand{\url}[1]{#1}
\csname url@rmstyle\endcsname
\providecommand{\newblock}{\relax}
\providecommand{\bibinfo}[2]{#2}
\providecommand\BIBentrySTDinterwordspacing{\spaceskip=0pt\relax}
\providecommand\BIBentryALTinterwordstretchfactor{4}
\providecommand\BIBentryALTinterwordspacing{\spaceskip=\fontdimen2\font plus
\BIBentryALTinterwordstretchfactor\fontdimen3\font minus \fontdimen4\font\relax}
\providecommand\BIBforeignlanguage[2]{{%
\expandafter\ifx\csname l@#1\endcsname\relax
\typeout{** WARNING: IEEEtran.bst: No hyphenation pattern has been}%
\typeout{** loaded for the language `#1'. Using the pattern for}%
\typeout{** the default language instead.}%
\else
\language=\csname l@#1\endcsname
\fi
#2}}

\bibitem{zhang_task_2024}
L.~Zhang, M.~Li, W.~Yang, and S.~Yang, ``Task {Allocation} in {Heterogeneous} {Multi}-{Robot} {Systems} {Based} on {Preference}-{Driven} {Hedonic} {Game},'' in \emph{2024 {IEEE} {International} {Conference} on {Robotics} and {Automation} ({ICRA})}, May 2024, pp. 8967--8972.

\bibitem{gosrich_multi-robot_2023}
W.~Gosrich, S.~Mayya, S.~Narayan, M.~Malencia, S.~Agarwal, and V.~Kumar, ``\BIBforeignlanguage{en}{Multi-{Robot} {Coordination} and {Cooperation} with {Task} {Precedence} {Relationships}},'' May 2023.

\bibitem{gombolay_fast_2018}
M.~C. Gombolay, R.~J. Wilcox, and J.~A. Shah, ``Fast {Scheduling} of {Robot} {Teams} {Performing} {Tasks} {With} {Temporospatial} {Constraints},'' \emph{IEEE Transactions on Robotics}, vol.~34, pp. 220--239, Feb. 2018.

\bibitem{ansari_colossi_2024}
I.~Ansari, A.~Mohammed, Y.~Ansari, M.~Yusuf~Ansari, S.~Razak, and E.~Feo~Flushing, ``{CoLoSSI}: {Multi}-{Robot} {Task} {Allocation} in {Spatially}-{Distributed} and {Communication} {Restricted} {Environments},'' \emph{IEEE Access}, vol.~12, 2024.

\bibitem{chakraa_optimization_2023}
H.~Chakraa, F.~Guérin, E.~Leclercq, and D.~Lefebvre, ``\BIBforeignlanguage{en}{Optimization techniques for {Multi}-{Robot} {Task} {Allocation} problems: {Review} on the state-of-the-art},'' \emph{\BIBforeignlanguage{en}{Robotics and Autonomous Systems}}, vol. 168, p. 104492, Oct. 2023.

\bibitem{bischoff_multi-robot_2020}
E.~Bischoff, F.~Meyer, J.~Inga, and S.~Hohmann, ``Multi-{Robot} {Task} {Allocation} and {Scheduling} {Considering} {Cooperative} {Tasks} and {Precedence} {Constraints},'' May 2020.

\bibitem{ramachandran_resilience_2019}
R.~K. Ramachandran, J.~A. Preiss, and G.~S. Sukhatme, ``Resilience by {Reconfiguration}: {Exploiting} {Heterogeneity} in {Robot} {Teams},'' in \emph{2019 {IEEE}/{RSJ} {International} {Conference} on {Intelligent} {Robots} and {Systems} ({IROS})}, Nov. 2019.

\bibitem{khamis_multi-robot_2015}
A.~Khamis, A.~Hussein, and A.~Elmogy, ``Multi-robot {Task} {Allocation}: {A} {Review} of the {State}-of-the-{Art},'' in \emph{Cooperative {Robots} and {Sensor} {Networks} 2015}, A.~Koubâa and J.~Martínez-de Dios, Eds.\hskip 1em plus 0.5em minus 0.4em\relax Cham: Springer International Publishing, 2015, pp. 31--51.

\bibitem{quinton_market_2023}
F.~Quinton, C.~Grand, and C.~Lesire, ``Market {Approaches} to the {Multi}-{Robot} {Task} {Allocation} {Problem}: a {Survey},'' \emph{Journal of Intelligent \& Robotic Systems}, vol. 107, no.~2, p.~29, Feb. 2023.

\bibitem{babincsak_ant_2023}
W.~Babincsak, A.~Aswale, and C.~Pinciroli, ``Ant {Colony} {Optimization} for {Heterogeneous} {Coalition} {Formation} and {Scheduling} with {Multi}-{Skilled} {Robots},'' in \emph{2023 {International} {Symposium} on {Multi}-{Robot} and {Multi}-{Agent} {Systems} ({MRS})}, Dec. 2023, pp. 121--127.

\bibitem{fu_robust_2021}
B.~Fu, W.~Smith, D.~Rizzo, M.~Castanier, M.~Ghaffari, and K.~Barton, ``Robust {Task} {Scheduling} for {Heterogeneous} {Robot} {Teams} under {Capability} {Uncertainty},'' \emph{IEEE Transactions on Robotics}, June 2021.

\bibitem{muhuri_immigrants_2017}
P.~Muhuri and A.~Rauniyar, ``Immigrants {Based} {Adaptive} {Genetic} {Algorithms} for {Task} {Allocation} in {Multi}-{Robot} {Systems},'' \emph{International Journal of Computational Intelligence and Applications}, vol.~16, p. 1750025, Dec. 2017.

\bibitem{gini_multi-robot_2017}
M.~Gini, ``Multi-{Robot} {Allocation} of {Tasks} with {Temporal} and {Ordering} {Constraints},'' \emph{Proceedings of the AAAI Conference on Artificial Intelligence}, vol.~31, no.~1, Feb. 2017.

\bibitem{gerkey_formal_2004}
B.~P. Gerkey and M.~J. Matarić, ``A {Formal} {Analysis} and {Taxonomy} of {Task} {Allocation} in {Multi}-{Robot} {Systems},'' \emph{The International Journal of Robotics Research}, vol.~23, no.~9, pp. 939--954, Sept. 2004.

\bibitem{korsah_comprehensive_2013}
G.~A. Korsah, A.~Stentz, and M.~B. Dias, ``A comprehensive taxonomy for multi-robot task allocation,'' \emph{The International Journal of Robotics Research}, vol.~32, no.~12, pp. 1495--1512, Oct. 2013.

\bibitem{aswale_heterogeneous_2023}
A.~Aswale and C.~Pinciroli, ``Heterogeneous {Coalition} {Formation} and {Scheduling} with {Multi}-{Skilled} {Robots},'' June 2023.

\bibitem{ansari_cooperative_2020}
I.~Ansari, A.~Mohamed, E.~F. Flushing, and S.~Razak, ``Cooperative and load-balancing auctions for heterogeneous multi-robot teams dealing with spatial and non-atomic tasks,'' in \emph{2020 {IEEE} {International} {Symposium} on {Safety}, {Security}, and {Rescue} {Robotics} ({SSRR})}, Nov. 2020.

\bibitem{irfan_auction-based_2016}
M.~Irfan and A.~Farooq, ``Auction-based task allocation scheme for dynamic coalition formations in limited robotic swarms with heterogeneous capabilities,'' in \emph{2016 {International} {Conference} on {Intelligent} {Systems} {Engineering} ({ICISE})}, Jan. 2016, pp. 210--215.

\bibitem{chakraa_centralized_2023}
H.~Chakraa, E.~Leclercq, F.~Guérin, and D.~Lefebvre, ``A {Centralized} {Task} {Allocation} {Algorithm} for a {Multi}-{Robot} {Inspection} {Mission} {With} {Sensing} {Specifications},'' \emph{IEEE Access}, vol.~11, 2023.

\bibitem{arif_robot_2021}
M.~U. Arif, ``Robot coalition formation against time-extended multi-robot tasks,'' \emph{International Journal of Intelligent Unmanned Systems}, vol.~10, pp. 468--481, June 2021.

\bibitem{liu_strength_2023}
X.-F. Liu, Y.~Fang, Z.-H. Zhan, and J.~Zhang, ``Strength {Learning} {Particle} {Swarm} {Optimization} for {Multiobjective} {Multirobot} {Task} {Scheduling},'' \emph{IEEE Transactions on Systems, Man, and Cybernetics: Systems}, vol.~53, no.~7, pp. 4052--4063, July 2023.

\bibitem{prorok_holy_2021}
A.~Prorok, J.~Blumenkamp, Q.~Li, R.~Kortvelesy, Z.~Liu, and E.~Stump, ``The {Holy} {Grail} of {Multi}-{Robot} {Planning}: {Learning} to {Generate} {Online}-{Scalable} {Solutions} from {Offline}-{Optimal} {Experts},'' July 2021.

\bibitem{wang_heterogeneous_2022}
Z.~Wang, C.~Liu, and M.~Gombolay, ``Heterogeneous graph attention networks for scalable multi-robot scheduling with temporospatial constraints,'' \emph{Autonomous Robots}, vol.~46, no.~1, pp. 249--268, Jan. 2022.

\bibitem{paul_learning_2022}
S.~Paul, P.~Ghassemi, and S.~Chowdhury, ``Learning {Scalable} {Policies} over {Graphs} for {Multi}-{Robot} {Task} {Allocation} using {Capsule} {Attention} {Networks},'' May 2022.

\bibitem{altundas_learning_2022}
B.~Altundas, Z.~Wang, J.~Bishop, and M.~Gombolay, ``Learning {Coordination} {Policies} over {Heterogeneous} {Graphs} for {Human}-{Robot} {Teams} via {Recurrent} {Neural} {Schedule} {Propagation},'' in \emph{2022 {IEEE}/{RSJ} {International} {Conference} on {Intelligent} {Robots} and {Systems} ({IROS})}, Oct. 2022.

\bibitem{deng_learning_2022}
F.~Deng, H.~Huang, L.~Fu, H.~Yue, J.~Zhang, Z.~Wu, and T.~L. Lam, ``A {Learning} {Approach} to {Multi}-robot {Task} {Allocation} with {Priority} {Constraints} and {Uncertainty},'' in \emph{2022 {IEEE} {International} {Conference} on {Industrial} {Technology} ({ICIT})}, Aug. 2022, pp. 1--8.

\bibitem{dai_dynamic_2024}
W.~Dai, A.~Bidwai, and G.~Sartoretti, ``Dynamic {Coalition} {Formation} and {Routing} for {Multirobot} {Task} {Allocation} via {Reinforcement} {Learning},'' in \emph{2024 {IEEE} {International} {Conference} on {Robotics} and {Automation} ({ICRA})}.\hskip 1em plus 0.5em minus 0.4em\relax Yokohama, Japan: IEEE, May 2024, pp. 16\,567--16\,573.

\bibitem{dai_heterogeneous_2025}
W.~Dai, U.~Rai, J.~Chiun, Y.~Cao, and G.~Sartoretti, ``Heterogeneous {Multi}-robot {Task} {Allocation} and {Scheduling} via {Reinforcement} {Learning},'' \emph{IEEE Robotics and Automation Letters}, vol.~10, no.~3, pp. 2654--2661, Mar. 2025.

\bibitem{gao_collaborative_2023}
P.~Gao, S.~Siva, A.~Micciche, and H.~Zhang, ``Collaborative {Scheduling} with {Adaptation} to {Failure} for {Heterogeneous} {Robot} {Teams},'' in \emph{2023 {IEEE} {International} {Conference} on {Robotics} and {Automation} ({ICRA})}, May 2023, pp. 1414--1420.

\bibitem{jose_learning_2024}
W.~J. Jose and H.~Zhang, ``Learning for {Dynamic} {Subteaming} and {Voluntary} {Waiting} in {Heterogeneous} {Multi}-{Robot} {Collaborative} {Scheduling},'' \emph{IEEE International Conference on Robotics and Automation (ICRA)}, 2024.

\bibitem{velickovic_graph_2018}
P.~Veličković, G.~Cucurull, A.~Casanova, A.~Romero, P.~Liò, and Y.~Bengio, ``Graph {Attention} {Networks},'' Feb. 2018.

\bibitem{vaswani_attention_2017}
A.~Vaswani, N.~Shazeer, N.~Parmar, J.~Uszkoreit, L.~Jones, A.~N. Gomez, L.~Kaiser, and I.~Polosukhin, ``Attention {Is} {All} {You} {Need},'' June 2017.

\bibitem{xu_empirical_2015}
B.~Xu, N.~Wang, T.~Chen, and M.~Li, ``Empirical {Evaluation} of {Rectified} {Activations} in {Convolutional} {Network},'' Nov. 2015.

\bibitem{kingma_adam_2017}
D.~P. Kingma and J.~Ba, ``Adam: {A} {Method} for {Stochastic} {Optimization},'' Jan. 2017.

\end{thebibliography}
 
\end{document}